%% file: main.tex
\documentclass{article}
\usepackage{ijcai21}

\usepackage{times}
\usepackage{soul}
\usepackage{url}
\usepackage[hidelinks]{hyperref}
\usepackage[utf8]{inputenc}
\usepackage[small]{caption}
\usepackage{graphicx}
\graphicspath{ {./fig/} }
\usepackage{amsmath}
\usepackage{amssymb}
\usepackage{amsfonts}
\usepackage{amsthm}
\usepackage{xcolor}
\usepackage{booktabs}
\usepackage{algorithm}
\usepackage{algorithmic}
\usepackage{stfloats}
\usepackage{multirow}
\usepackage{xspace}
\usepackage{hyperref}
\usepackage{setspace}
\usepackage{wrapfig}
\newcommand{\ie}{\textit{i}.\textit{e}.\xspace}
\newcommand{\eg}{\textit{e}.\textit{g}.\xspace}
\newcommand{\vct}[1]{\boldsymbol{#1}} 
\newcommand{\mat}[1]{\boldsymbol{#1}} 
\newcommand\inv[1]{#1\raisebox{1.15ex}{$\scriptscriptstyle-\!1$}}

\newcommand{\methodname}{AgeFlow\xspace}

\title{\methodname: Conditional Age Progression and Regression with Normalizing Flows}
\author{
Zhizhong Huang$^1$\qquad
Shouzhen Chen$^1$\qquad
Junping Zhang$^1$\qquad
Hongming Shan$^{2,3}$\thanks{Corresponding author}\\
\affiliations
$^1$Shanghai Key Lab of Intelligent Information Processing, School of Computer Science\\
Fudan University, Shanghai 200433, China\\
$^{2}$Institute of Science and Technology for Brain-inspired Intelligence and MOE Frontiers Center \\for Brain Science,  Fudan University, Shanghai 200433, China\\
$^{3}$Shanghai Center for Brain Science and Brain-inspired Technology, Shanghai 200031, China\\
\emails
\{zzhuang19, chensz19, jpzhang, hmshan\}@fudan.edu.cn
}

\begin{document}

\maketitle

\begin{abstract}

Age progression and regression aim to synthesize photorealistic appearance of a given face image with aging and rejuvenation effects, respectively. Existing generative adversarial networks (GANs) based methods suffer from the following three major issues: 1) unstable training introducing strong ghost artifacts in the generated faces, 2) unpaired training leading to unexpected changes in facial attributes such as genders and races, and 3) non-bijective age mappings increasing the uncertainty in the face transformation. To overcome these issues, this paper proposes a novel framework, termed \methodname, to integrate the advantages of both flow-based models and GANs. The proposed \methodname contains three parts: an encoder that maps a given face to a latent space through an invertible neural network, a novel invertible conditional translation module (ICTM) that translates the source latent vector to target one, and a decoder that reconstructs the generated face from the target latent vector using the same encoder network; all parts are invertible achieving bijective age mappings. The novelties of ICTM are two-fold. First, we propose an attribute-aware knowledge distillation to learn the manipulation direction of age progression while keeping other unrelated attributes unchanged, alleviating unexpected changes in facial attributes. Second,  we propose to use GANs in the latent space to ensure the learned latent vector indistinguishable from the real ones, which is much easier than traditional use of GANs in the image domain. Experimental results demonstrate superior performance over existing GANs-based methods on two benchmarked datasets. The source code is available at~\url{https://github.com/Hzzone/AgeFlow}.

\end{abstract}

\input{intro.tex}
\input{method.tex}
\input{exp.tex}

\input{conc}

\clearpage

\bibliographystyle{named}

\input{appendix}

\end{document}

%% file: intro.tex

\section{Introduction}

Age progression and regression, also known as face aging and rejuvenation, are to synthesize the appearance of a given face image at different ages with natural aging/rejuvenation effects. Despite their broad applications ranging from entertainment to social security~\cite{park2010age,huang2021age}, the intrinsic complexities such as complicated translation patterns, various expressions, and diversified races pose challenges to existing methods. Recent studies~\cite{li2020hierarchical,or2020lifespan,huang2021routinggan} resort to generative adversarial networks~(GANs)~\cite{goodfellow2014generative} and its variant conditional GANs~(cGANs)~\cite{mirza2014conditional} to take advantage of massive unpaired age data, achieving more impressive results than conventional methods~\cite{wang2016recurrent}. 

\begin{figure}
    \centering
    \includegraphics[width=1\linewidth]{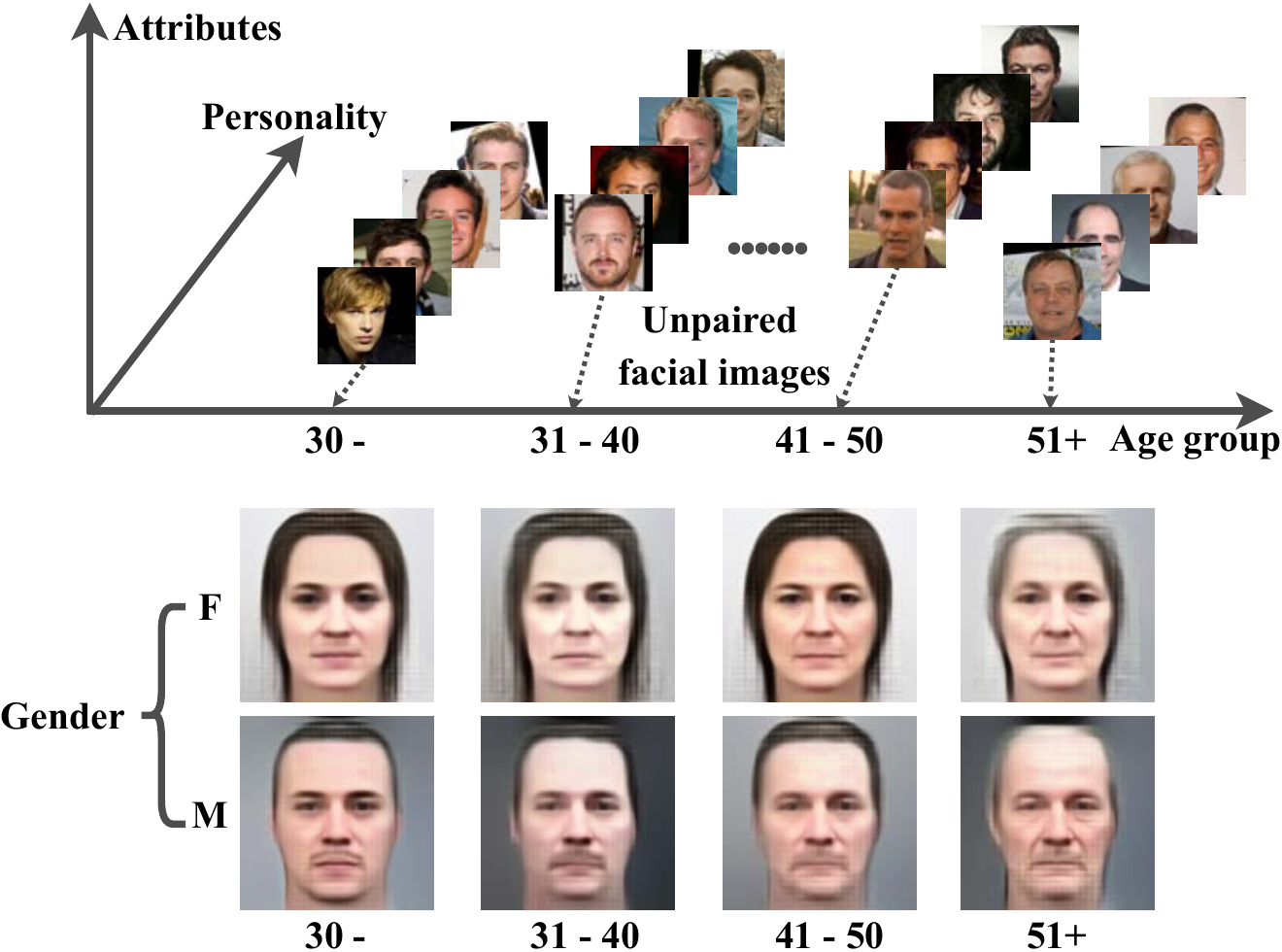}
     \vspace{-5mm}
    \caption{
        Top: Illustration of a face latent space learned by a flow-based generative model, in which faces align with different facial conditions. Bottom: The faces reconstructed from the \emph{average} latent variables for each attribute pair, \eg genders and age groups.
    }
    \label{fig:manifold}
\end{figure}

In spite of significant progress in meeting three important requirements of face aging/rejuvenation---image quality, age accuracy, and identity preservation, these GANs-based methods suffer the following shortcomings. \textbf{First}, the unstable training of GANs introduces strong ghost artifacts in the generated faces, which severely compromises the image quality of age progression/regression. \textbf{Second}, the unpaired training of GANs in the image domain may lead to unexpected changes in facial attributes due to the neglect of some important conditional information of the input~(\eg, gender and race), thus challenging the identity preservation of input faces. \textbf{Third}, these GANs-based methods handle all age mappings together with one single network under the control of an age condition, increasing the uncertainty in the face transformation with unsatisfied age accuracy. Several methods have been proposed to address these problems. Initial attempts~\cite{yang2018learning,liu2019attribute,huang2021pfa} mainly focus on training several independent models for each age mapping. Considering the different patterns of age progression and regression, \cite{song2018dual} used two separate generators to model both age changing processes separately, and \cite{li2020age} further extended it with a spatial attention mechanism. However, they usually suffer from considerably computational costs and are limited with the age label as prior.

Inspired by the great success of flow-based generative models~\cite{dinh2014nice,dinh2016density,kingma2018glow} in exact latent-variable inference, this paper exploits the advantages of flow-based generative models and GANs for both age progression/regression. Most importantly, a more stable training process can be achieved by the flow-based generative models as they are directly optimized by the log-likelihood objective function instead of an adversarial competition. Figure~\ref{fig:manifold} (top) illustrates that the faces in the latent space align with the facial attributes, personality, and ages. Therefore, the average latent variables for each attribute can be computed as the direction for manipulating the attribute-aware latent space. In this paper, we mainly focus on manipulating the ages of faces while keeping other attributes unchanged for the present age progression and regression tasks. Therefore, an average latent vector for each attribute pair can keep the attribute consistent during face aging/rejuvenation. Taking the genders as an example, Figure~\ref{fig:manifold} (bottom) shows the reconstructed prototypes, \ie, average latent vectors of different age groups and genders. Remarkably, the wrinkle appears, hair becomes white, and beard grows \emph{only for male} on the prototype faces as expected when faces age from left to right. However, the original flow-based models lack the capability of conditionally facial manipulation. To achieve conditional age progression/regression similar to the cGANs-based method, the learned manipulation direction can be used to guide the generative models in a knowledge distilling manner with the advantage of GANs to ensure the learned latent vector indistinguishable from the real one.

\begin{figure*}[t!]
    \centering
    \includegraphics[width=1.0\linewidth]{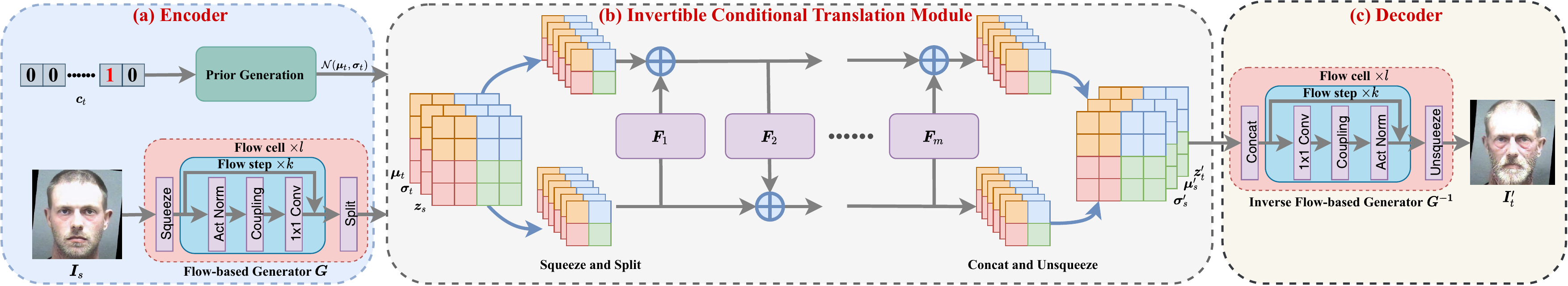}
    \vspace{-3mm}
    \caption{
        The overall framework of the proposed \methodname. (a) The encoder maps an input image $\mat{I}_s$ into $\vct{z}_s$ in the latent space using a flow-based generator $G$, and a prior generation module converts one-hot encoding of target age $\vct{c}_t$ to a Gaussian distribution, $\mathcal{N}(\vct{\mu}_t, \vct{\sigma}_t^2)$. (b) The invertible conditional translation module (ICTM) takes $\vct{z}_s$, $\vct{\mu}_t$ and $\vct{\sigma}_t$ as input and achieves conditional age progression/regression with normalizing flows in the latent space, outputting the progressed/regressed latent variables $\vct{z}_t^{\prime}$ and source age distribution from $\mat{I}_s$ to keep the cycle consistency. ICTM contains $m$ flows, each of which contains two convolutional layers, one channel attention module, and one convolutional layer initialized by zero. (c) The inverse function of $G$ reconstructs the generated face $\mat{I}_t^{\prime}$ from $\vct{z}_t^{\prime}$.
        }
    \label{fig:inference}
\end{figure*}

\paragraph{\textbf{Contribution}}\quad The contributions are summarized as follows. 1) We propose a novel framework, termed \methodname, to integrate the advantages of both flow-based models and GANs, which contains three parts---encoder, invertible conditional translation module (ICTM), and decoder; all are invertible reducing the uncertainty in face transformation. 2) We introduce an attribute-aware knowledge distillation to learn the manipulation direction of face progression while keeping other unrelated attributes unchanged, which can reduce unexpected changes in facial attributes. 3) We apply GANs in the latent space to ensure the learned latent vector indistinguishable from the real ones, which is much easier than the traditional use of GANs in the image domain.

%% file: method.tex
\section{Methodology}\label{sec:method}

\subsection{Problem Formulation}

Given a face image $\mat{I}_s$ at source age $\vct{c}_s$, age progression/regression aims to learn a neural network to naturally translate $\mat{I}_s$ into $\mat{I}_t$ conditioned on a given target age $\vct{c}_t$. Formally, the age progression/regression process of the cGANs-based methods can be formulated as:
\begin{align}\label{Eq:traditional}
    \mat{I}_t^{\prime} = G(\mat{I}_s, \vct{c}_t),
\end{align}
where $\mat{I}_t^{\prime}$ represents the generated face from $\mat{I}_s$.
However, the translation  $G:\mat{I}_{s} \to \mat{I}_{t}$ is highly under-constrained due to the unpaired training of GANs, misleading the model to learn the patterns other than face aging/rejuvenation. One possible solution is to regularize the mappings with cycle-consistency~\cite{zhu2017unpaired}, which learns backwards translation $\mat{I}_{t} \to \mat{I}_{s}$ using a separate model~\cite{song2018dual,li2020age} to approximately reconstruct $\mat{I}_{s}$. However, the cycle-consistency brings more uncertainty of age mappings and cannot guarantee the bijective age mapping, especially when the gap between $\vct{c}_s$ and $\vct{c}_t$ becomes large.

To address this problem, we propose to model the age progression/regression as one invertible process using an invertible neural network~\cite{dinh2016density}. Intuitively, the source age condition $\vct{c}_t$ could be extracted from the input face during performing the age progression/regression with the target age condition $\vct{c}_t$. In this way, the flexibility of cGANs-based methods can be still enjoyed.
Therefore, the invertible age progression/regression process from the source age group $s$ to target age group $t$ can be formulated as follows:
\begin{align}
    \mat{I}_t^{\prime}, \vct{c}_s^{\prime} = G(\mat{I}_s, \vct{c}_t).
\end{align}

The inference phase of \methodname is illustrated in Figure~\ref{fig:inference}. Given an input face $\mat{I}_s$, instead of directly modeling the patterns in the pixel space, the proposed method achieves age progression/regression on the latent vectors $\vct{z}=G(\mat{I})$ due to its reversibility, where $G$ is a trained generative flow-based model~\cite{kingma2018glow}. Furthermore, we
use a prior generator to produce the distribution of age condition as the conventional one-hot condition in the context of cGANs-based methods to keep consistency. This is because we found that the distribution contains more information than simple one-hot encoding that is too restrictive for the model to learn. In summary, the invertible conditional translation can be formulated as:
\begin{align}\label{eq_ictm}
    \vct{z}_t^{\prime}, \vct{\mu}_s^{\prime}, \vct{\sigma}_s^{\prime} = T(\vct{z}_s, \vct{\mu}_t, \vct{\sigma}_t),
\end{align}
where $T$ is the proposed invertible conditional translation module, and $\vct{\mu}$ and $\vct{\sigma}$ are the parameters of the Gaussian distribution from the prior generator for the age condition. Note that the decoder can easily reconstruct the learned latent vector back to the image domain through the inverse of the same encoder network, \ie, $\inv{G}$. In other words, the cycle-consistency is achieved by the invertible network rather than reconstructing the input faces by another model.

\subsection{Network Architecture}

\begin{figure}[t!]
    \centering
    \includegraphics[width=1\linewidth]{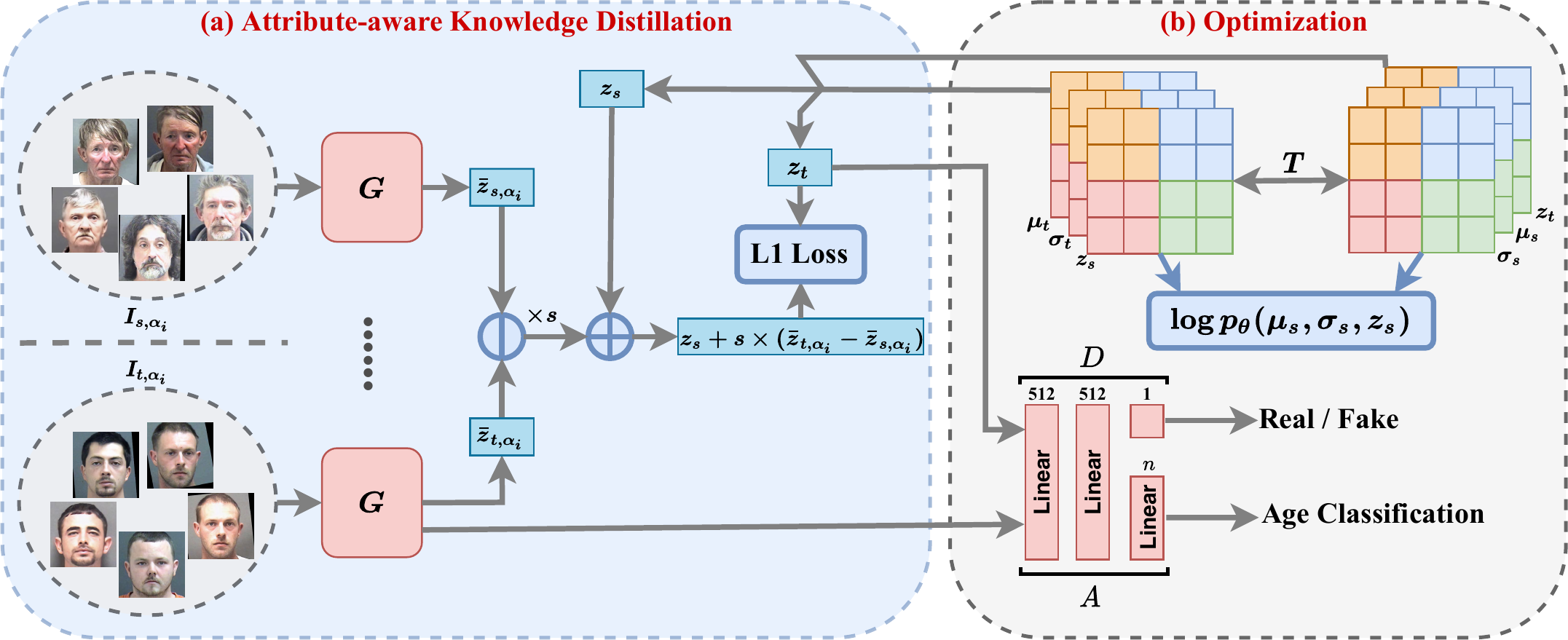}
    \caption{
        Training illustration of the ICTM. The average vectors $\vct{z}_{i, \alpha_i}$ for each attribute pair are first computed from a set of images to form the manipulation direction, where $i$ and $\alpha_i$  denotes the age group and facial attribute, respectively. They then become the label of ICTM to guide the directions of age progression/regression, \ie, distilling the knowledge of a GLOW model learned from the dataset to ICTM. There also includes GANs and age classification task for better image quality and age accuracy with the maximum log-likelihood estimation to keep the cycle consistency. 
        }
    \label{fig:training}
\end{figure}

As shown in Figures~\ref{fig:inference} and~\ref{fig:training}, our \methodname consists of three key components to achieve an invertible conditional age progression/regression:  1) a flow-based encoder $G$ that maps the input images into latent space and the decoder that is inverse function $\inv{G}$; 2) an invertible conditional translation module $T$ that achieves age progression/regression in the latent space; and 3) a discriminator to force $T$ to generate age progressed/regressed latent vectors indistinguishable from real samples in the GAN framework~\cite{goodfellow2014generative} taking advantage of massive unpaired age data. Here we detail the network architectures of these three components.

\paragraph{\textbf{Flow-based Generator}}\quad
Flow-based generative models or GLOW~\cite{kingma2018glow} typically set up a bijective mapping between the image and latent space with invertible networks, which formulates the encoding process as $\vct{z}=G(\mat{I})$ and the generation one as the reverse procedure $\mat{I}=\inv{G}(\vct{z})$, where $\vct{z}$ has a prior Gaussian distribution $p(\vct{z})$ and $G$ is the invertible encoder. Due to the optimization of exact log-likelihood objective, flow-based models provide direct interpolations between data points and meaningful modifications of existing data points; in contrast, there is no encoder in GANs to model the full data distribution while VAEs maximize the evidence lower bound but generate blurry images. Here we adopt the similar structure of GLOW~\cite{kingma2018glow} for modeling the latent space with a standard Gaussian distribution. Specifically, as shown in Figure~\ref{fig:inference}, there are $l\times k$ normalizing flows, each of which includes ActNorm, Addictive Coupling, and invertible $1\times1$ convolutional layers, where the number of filters increases twice and the size of feature maps decreases half for every $k$ flows at each $l$ scale. In this paper, we adopt $6\times 32$ for input images with the size of $256\times 256$.  Therefore, the loss function (maximum likelihood estimation) to optimize $G$ can be written as:
\begin{align}
    \mathcal{L}_{G}=-\mathbb{E}_{\vct{z} \sim p_\theta(\vct{z})}\left[\log p_\theta(\vct{z})+\log \left|\operatorname{det} \frac{\mathrm{d} G}{\mathrm{d} \mat{I}}\right|\right].
\end{align}
Once trained, $G$ is fixed to extract the latent vectors and the inverse of the trained $G$, $\inv{G}$, also serves as the decoder that maps the latent vector back to the image domain. In doing so, we can only train a small invertible translation network in the latent space to achieve age progression/regression, avoiding considerably computational cost of directly training invertible network from source age to target one in the image domain.

\paragraph{\textbf{Invertible Conditional Translation Module}}\quad
Given a latent vector $\vct{z}_s$ from the source age group $\vct{c}_s$, the age progression/regression is to translate $\vct{z}_s$ to $\vct{z}_t$ at the target age $\vct{c}_t$. As illustrated in Figure~\ref{fig:inference}, the proposed invertible conditional translation module~(ICTM) $T$ leverages the invertible network to achieve both age progression and regression, where the cycle consistency can be easily achieved due to its reversibility. In a sense, the inverse function of $T$ plays a similar role in~\cite{song2018dual,li2020age} that achieves age regression independent of age progression. To further keep the flexibility of the cGANs-based method that achieves age progression/regression with only changing the target age condition, the translation module concatenates the latent vector $\vct{z}_s$ with the distributions of target age condition estimated from the prior generator to achieve conditional translation. The reason why we adopt the distribution of age conditions is that the one-hot encoding used in previous works is too restrictive for the model to keep such cycle consistency and extract the age information from the input faces. Specifically, ICTM consists of $m$ flows similar to previous invertible networks~\cite{dinh2016density}. Each flow contains two convolutional layers, a channel attention module to enable the model to focus on changing the necessary latent variable, and a convolutional layer whose weights are initialized by zero to perform an identity function at the beginning of training. Finally, the progressed/regressed latent vectors and distributions of source age condition can be reconstructed from the two-way outputs of the ICTM, as indicated in Eq.~\eqref{eq_ictm}.

\paragraph{\textbf{Discriminator}}\quad
We adopt a simple multi-layer perceptron (MLP) with two linear layers of 512 neurons as the discriminator $D$ in our framework, each of which is followed by a spectral normalization~\cite{miyato2018spectral} and a leaky ReLU activation whose negative slope is 0.2. At the bottom, we also append a linear layer with 1 neuron to output the confidence of GANs and another linear layer of $n$ neurons for age classification to improve the age accuracy of our method, where $n$ is the number of age groups.

\subsection{Loss Function}

The overall loss function to optimize the proposed ICTM contains four components for age progression/regression: 1) \textbf{Attribute-aware Knowledge Distillation Loss} aims to distill the manipulation direction of latent vectors for the ICTM, which can preserve the flexibility of original cGANs-based methods with the facial attributes unchanged; 2) \textbf{Adversarial Loss} intends to force the ICTM to produce progressed/regressed latent variables indistinguishable from real ones; 3) \textbf{Age Classification Loss} expects to improve the aging accuracy, and 4) \textbf{Consistency Loss} seeks to regularize the ICTM with the cycle consistency.

\paragraph{\textbf{Attribute-aware Knowledge Distillation Loss}}\quad
The flow-based model can directly interpolate at the learned latent space and modify the data points to generate photorealistic images. For example, it can achieve image translation by applying the manipulation direction to the latent vector $\vct{z}_s$ of one given image, \ie adding an average vector computed from a few images of the binary classes, which can be formulated as follows:
\begin{align}\label{eq:manipulation}
\vct{z}_{t}^{\prime}=\vct{z}_s + s \times (\vct{z}_{\mathrm{pos}} - \vct{z}_{\mathrm{neg}}),
\end{align}
where $s$ is a scale factor controlling the degree of manipulating $\vct{z}_s$, and $\vct{z}_{\mathrm{pos}}$ and $\vct{z}_{\mathrm{neg}}$ are the average latent vectors computed from the sets of positive and negative samples, respectively. However, the original GLOW is not directly applicable to age progression/regression for the following reasons: 1) the facial attributes should be preserved during age progression/regression and 2) it cannot achieve conditional age progression/regression without the age label as prior and fails to ensure satisfied age accuracy; see the experiments for more evidence. To push the ICTM to manipulate the latent variables towards the direction of age progression/regression while keeping other attributes unchanged, we propose to use the learned manipulation direction to guide the ICTM in a knowledge distillation manner. Figure~\ref{fig:training}(a) shows the details of proposed attribute-aware knowledge distillation. Specifically, to preserve the facial attributes, we first compute the average vectors for each attribute pair from a set of images. By replacing the average latent vectors in Eq.~\eqref{eq:manipulation}, the consistency of facial attributes can be easily achieved. In addition, the learned direction can be further used as the supervision signals of ICTM optimized by the L1 distance. In summary, the knowledge distillation loss can be written as:
\begin{align}
    \mathcal{L}_{\mathrm{akd}} = \mathbb{E}_{\vct{z}_s}\Big|\vct{z}_t^{\prime} - \big(\vct{z}_s + s \times (\bar{\vct{z}}_{t, \alpha_i} - \bar{\vct{z}}_{t, \alpha_i})\big)\Big|,
\end{align}
where $s$ is a hyperparameter to achieve better age accuracy, and $\alpha_i$ represents one attribute of $z_s$.

\paragraph{\textbf{Adversarial Loss}}\quad
In this paper, we employ least squares GANs~\cite{mao2017least} to optimize our discriminator in the latent space due to its stability, which can be defined as follows:
\begin{align}
    \mathcal{L}_{\mathrm{al}}=\frac{1}{2} \mathbb{E}_{\vct{z}_{s}}\left[\left(D\left(\vct{z}_t^{\prime}\right)-1\right)^{2}\right].
\end{align}

\paragraph{\textbf{Age Classification Loss}}\quad
To improve the age accuracy of the generated faces, we also include an age classification loss in the final loss function, which can be defined as:
\begin{equation}
    \mathcal{L}_{\mathrm{acl}}=\mathbb{E}_{\vct{z}_s}\Big[\ell\big(A(\vct{z}_t^{\prime}), t\big)\Big],
\end{equation}
where $\ell$ is the cross-entropy loss and $A$ denotes the age classifier as shown in Figure~\ref{fig:training}.

\paragraph{\textbf{Consistency Loss}}\quad
To keep the cycle consistency, the ICTM is supposed to output meaningful distribution of source age condition extracted from the input image. Therefore, we adopt a consistency loss that uses a standard Gaussian distribution to achieve the maximum log-likelihood estimation, which is formulated as:
\begin{align}
    \mathcal{L}_{\mathrm{cl}}=-\mathbb{E}_{\vct{z}_s \sim p_\theta(\vct{z}_s)}\left[\log p_\theta(\vct{\mu}_s^{\prime}, \vct{\sigma}_s^{\prime}, \vct{z}_s)\right].
\end{align}
Due to the reversibility of invertible networks, we only need to maintain such cycle consistency for the output of ICTM. Besides, the input $\vct{z}_s$ can be easily reconstructed without any information loss.

\paragraph{\textbf{Final Loss}}\quad
In summary, the age progression/regression process requires age progressed/regressed latent variable to meet the following requirements: 1) indistinguishable from the real one and thus can be decoded to a photo-realistic face; 2) belongs to the target age group; and 3) has the same identity as the input one while keeping other facial attributes unchanged. Therefore, the final loss function to optimize ICTM is expressed as:
\begin{align}
    \mathcal{L}_{T}=\lambda_{\mathrm{akd}} \mathcal{L}_{\mathrm{akd}} + \lambda_{\mathrm{al}} \mathcal{L}_{\mathrm{al}} + \lambda_{\mathrm{acl}} \mathcal{L}_{\mathrm{acl}} + \lambda_{\mathrm{cl}} \mathcal{L}_{\mathrm{cl}}.
\end{align}

The loss function to optimize discriminator $D$ is defined as:
\begin{align}
    \mathcal{L}_{D} =& \tfrac{1}{2} \mathbb{E}_{\vct{z}_t} (D(\vct{z}_t)-1)^{2} + \tfrac{1}{2} \mathbb{E}_{\vct{z}_s} D(\vct{z}_t^{\prime})^{2} + \lambda_{\mathrm{acl}}^D \mathcal{L}_{\mathrm{acl}},
\end{align}
where $\vct{z}_{t}^{\prime}$ is the learned latent vector belonging to target age group.
During the training, the ICTM and discriminator $D$ are updated alternatively.

%% file: exp.tex
\section{Experiments}\label{sec:exp}

\subsection{Implementation Details}

\paragraph{\textbf{Data Collection}}\quad
We conducted experiments on two benchmark age datasets: \textbf{MORPH}~\cite{ricanek2006morph} and \textbf{CACD}~\cite{chen2015face}. We also adopted \textbf{FG-NET} and \textbf{CelebA}~\cite{liu2015faceattributes} as external testing sets to validate the model trained with CACD for a fair comparison. All face images were cropped and aligned based on landmarks detected by MTCNN~\cite{zhang2016joint} with an image size of $256\times 256$. We randomly divided the dataset into two parts without identities overlapping: 80\% for training and the remaining for testing. Following the mainstream works~\cite{yang2018learning,liu2019attribute,li2020age,huang2021pfa}, we divided the face images into $n=4$ age groups, \ie, 30-, 31-40, 41-50, 50+.

\paragraph{\textbf{Training Details}}\quad
All models are implemented by PyTorch and trained by Adam optimizer with a fixed learning rate of $10^{-5}$ for the GLOW model and ICTM, and $10^{-4}$ for the discriminator. In addition, due to the limited GPU memory, we trained all models with a batch size of 16 on 4 NVIDIA V100 GPUs and the parameters are updated for every 4 iterations, equal to a batch size of 64. We first trained the GLOW model on CelebA~\cite{liu2015faceattributes} thanks to its diversity and a huge amount of face images for 1M iterations and then fine-tuned the models on each dataset with only 50K iterations. ICTM contains $m=32$ flows. The hyperparameters in the final loss are empirically set as follows: $\lambda_{\mathrm{al}}$ was  $1$; $\lambda_{\mathrm{cl}}$ was 0.01; $\lambda_{\mathrm{acl}}$ was $1$; $\lambda_{\mathrm{acl}}^D$ was $0.1$; and $\lambda_{\mathrm{akl}}$ was $1$. The $s$ in the knowledge distilling loss is set as 1.4 for MORPH and 1.8 for CACD. Here we emphasize that, to our best knowledge, we are the first work that has successfully reproduced GLOW from scratch with a high resolution of $256\times 256$. We also highlight that the pre-trained model will be very helpful for future works in the research community; see Appendix for more qualitative results of our trained GLOW model.

\subsection{Qualitative Results}\quad
\begin{figure}[t!]
    \centering
    \includegraphics[width=1\linewidth]{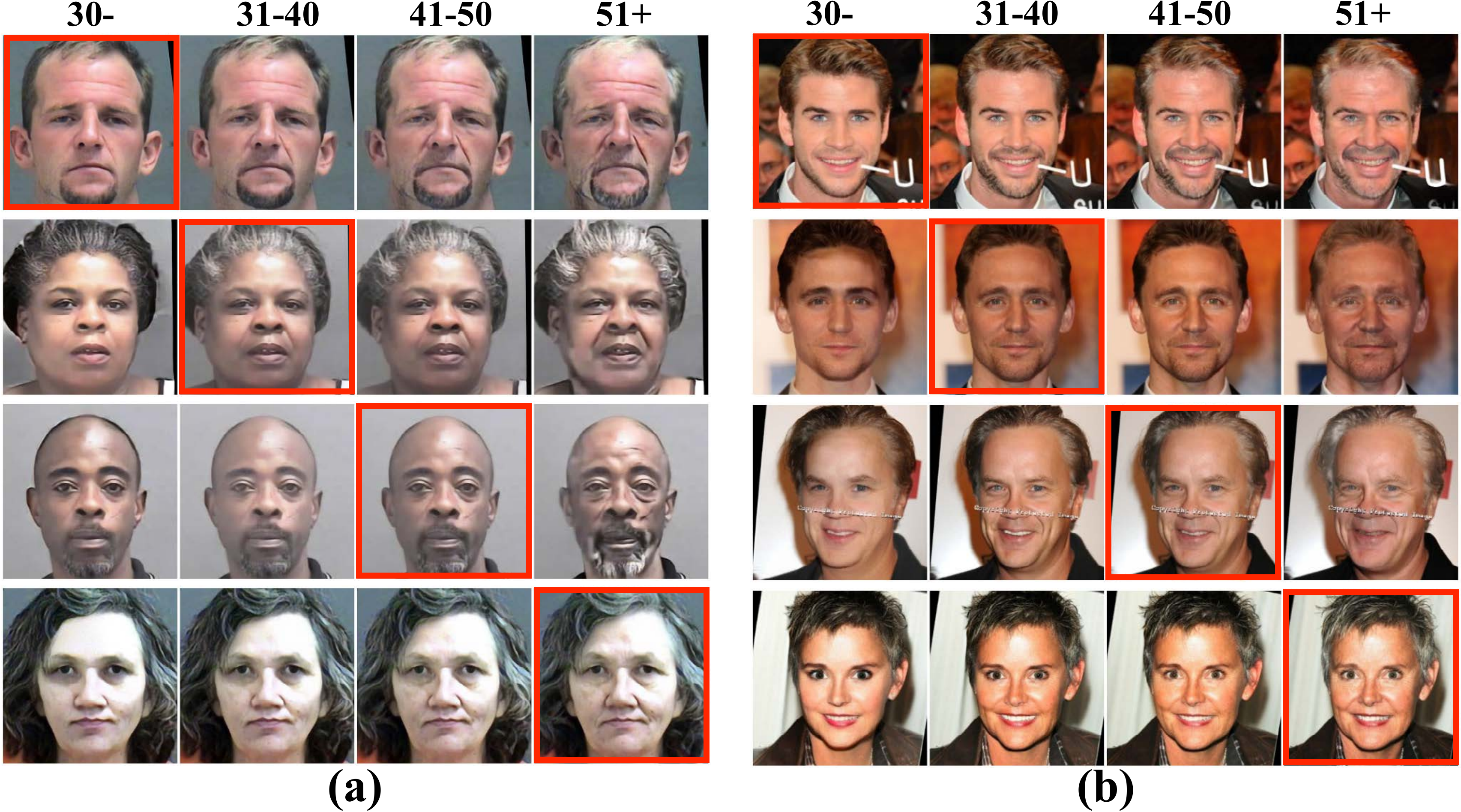}
    \caption{The generated faces by \methodname on (a) MORPH and (b) CACD. Red boxes indicate input faces for progression/regression.}
    \label{fig:exmaples}
\end{figure}
Figure~\ref{fig:exmaples} shows some generated faces by our \methodname from different ages to the other three age groups. Remarkably, the generated faces are photo-realistic and rendered with natural aging/rejuvenation effects; \eg, the skins are tightened at young ages while the hair becomes white at old ages. Most importantly, all faces present the same identities as the input faces, indicating that the identity consistency has been well preserved; see supplementary materials for more sample results on the FG-NET and CelebA datasets to demonstrate the generalization ability of the proposed \methodname.

\begin{figure*}[h!]
    \centering
    \includegraphics[width=.9\linewidth]{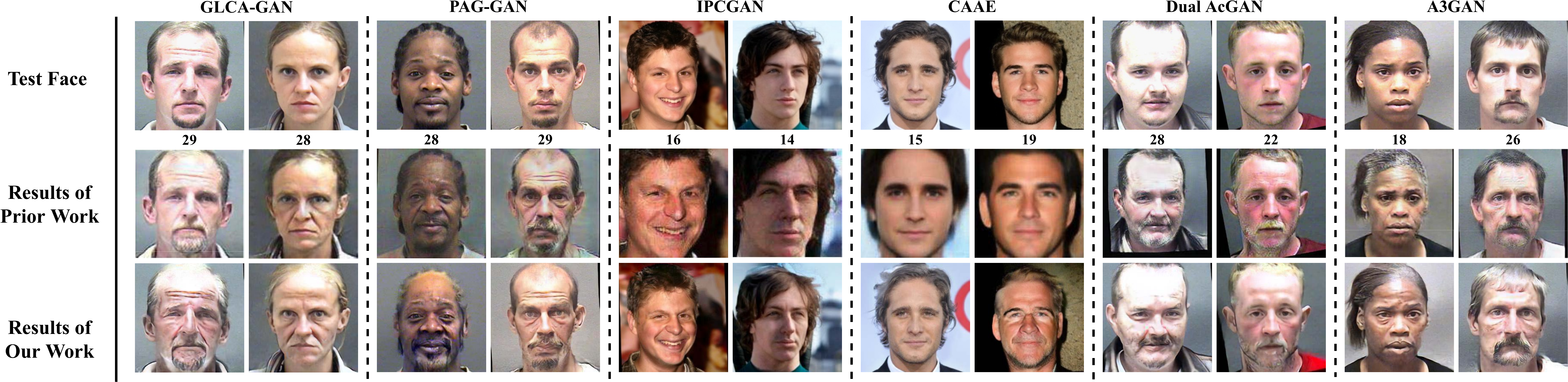}
    \caption{
        Performance comparison with prior work on the MORPH and CACD datasets. The three rows are the input young faces with their real age labels underneath, the results of prior work, and our results in the same age groups (51+), respectively.
    }
    \label{fig:method_comparision}
\end{figure*}

Figure~\ref{fig:method_comparision} shows the quantitative comparison between \methodname and other recent works including CAAE~\cite{zhang2017age}, IPCGAN~\cite{wang2018face}, PAG-GAN~\cite{yang2018learning}, GLCA-GAN~\cite{li2018global}, Dual AcGAN~\cite{li2020age}, and A3GAN~\cite{liu2019attribute}. Markedly, our results have the highest image quality with natural effects and fewer artifacts. This is a direct result of the combination of flow-based models and GANs. Specifically, although the GANs-based method can learn good translation patterns, it suffers from unstable training and mode collapse, which results in strong artifacts and blurring in the generated images. However, the flow-based models optimize the exact log-likelihood objective function and thus are better at learning the underlying distribution of the dataset. In addition, thanks to the reversibility of flow-based models, the ghost artifacts would, as expected, not appear as we only perform GANs in the latent space instead of the image space. We highlight that the results of baseline methods are directly referred from their published papers to avoid any bias resulting from our self-implementation.

\subsection{Quantitative Evaluations}

\begin{table*}[ht]
    \centering
    \caption{Quantitative results of different methods on MORPH and CACD datasets. The best results are shown in bold.}
    \vspace{-3mm}
    \begin{spacing}{1.0}
        \begin{tabular}{lrrrrclrrrr}
            \toprule
            &\multicolumn{4}{c}{\textbf{MORPH}} &&& \multicolumn{4}{c}{\textbf{CACD}} \\ 
            \cmidrule{1-5} \cmidrule{7-11} 
            Age group & 30 - & 31 - 40 & 41 - 50 & 50 + && Age group & 30 - & 31 - 40 & 41 - 50 & 50 +\\
            \midrule
            &\multicolumn{4}{c}{Age Accuracy~(\%)} & & &\multicolumn{4}{c}{Age Accuracy~(\%)} \\
            \cmidrule{1-5} \cmidrule{7-11} 
            CAAE      & 56.71 & 56.17 & 45.58 & 48.48 & & CAAE       & 57.50 & 42.60 & 50.51 & 24.97\\
            IPCGAN    & 75.71 & 80.30 & 48.91 & 53.71 & & IPCGAN     & 71.84 & 52.99 & 60.07 & 77.61\\
            WGLCA-GAN & 80.19  & 77.10  & 81.12  & 85.38 & & WGLCA-GAN  & 78.05  & 72.48  & 62.54  & 81.48\\
            \methodname (ours)   & \textbf{87.19}  & \textbf{91.85}  & \textbf{90.68}  & \textbf{90.71}  & & \methodname (ours)    & \textbf{86.49}  & \textbf{78.06}  & \textbf{85.91}  & \textbf{88.15}\\
            \cmidrule{1-5} \cmidrule{7-11} 
            &\multicolumn{4}{c}{Cosine Similarity} & & &\multicolumn{4}{c}{Cosine Similarity} \\
            \cmidrule{1-5} \cmidrule{7-11} 
            CAAE      & 0.265 & 0.289 & 0.279 & 0.220& & CAAE & 0.092 & 0.101 & 0.091 & 0.075\\
            IPCGAN    & 0.567 & 0.609 & 0.610 & 0.467& & IPCGAN & 0.445 & 0.554 & 0.569 & 0.462\\
            WGLCA-GAN & 0.671 & 0.792 & 0.764 & 0.614& & WGLCA-GAN & 0.752 & 0.803 & 0.769 & 0.758\\
            \methodname (ours)   & \textbf{0.897} & \textbf{0.923} & \textbf{0.903} & \textbf{0.767}& & \methodname (ours) & \textbf{0.905} & \textbf{0.926} & \textbf{0.919} & \textbf{0.847}\\ 
            \bottomrule
            \end{tabular}
    \end{spacing}
    \label{tab:performance}
\end{table*}

\begin{table*}[t!]
    \centering
    \caption{Attribute preservation for different methods on MORPH and CACD datasets. The best results are shown in bold.}
    \vspace{-3mm}
    \begin{tabular}{lcrrcrrrlcrr}
    \toprule
    & \multicolumn{7}{c}{\textbf{Gender}} && \multicolumn{3}{c}{\textbf{Race}} \\ 
    \cmidrule{2-8} \cmidrule{10-12} 
    & \multicolumn{3}{c}{MORPH} & \multicolumn{1}{l}{} & \multicolumn{3}{c}{CACD}  && \multicolumn{3}{c}{MORPH} \\
    \cmidrule{2-4} \cmidrule{6-8} \cmidrule{10-12} 
    Age Group & 31 - 40 & 41 - 50 & 51 +  & & 31 - 40 & 41 - 50 & 51 +  & & 31 - 40 & 41 - 50 & 51 + \\ 
    \midrule
    CAAE     & 94.53 & 94.59 & 93.37 & & 92.96 & 89.11 & 83.78 & & 97.01 & 96.73 & 93.84\\ 
    IPCGAN   & 97.30 & 97.48 & 95.14 & & 94.96 & 97.42 & 95.49 & & 96.87 & 97.10 & 97.05\\ 
    WGLCA-GAN& 97.71 & 98.95 & 96.20 & & 98.14 & 95.64 & 95.69 & & 98.53 & 98.26 & 96.28\\ 
    \methodname (ours)  & \textbf{98.19} & \textbf{99.26} & \textbf{98.21} & & \textbf{98.63} & \textbf{98.02} & \textbf{98.49} & & \textbf{98.85} & \textbf{98.80} & \textbf{97.72}\\ 
    \bottomrule
    \end{tabular}
    \label{tab:attribute_performance}
\end{table*}

We conduct qualitative comparisons with prior work including CAAE~\cite{zhang2017age}, IPCGAN~\cite{wang2018face}, and WGLCA-GAN~\cite{li2019global} with an ablation study. We employ two widely-used quantitative metrics to evaluate age progression/regression methods:  age accuracy and identity preservation, \ie, the progressed/regressed faces should ideally belong to the target age group with identities preserved. We also evaluated our method in preserving the facial attributes of input faces to demonstrate the effectiveness of the proposed attribute-aware knowledge distillation loss.

\paragraph{\textbf{Age Accuracy}}\quad
Following~\cite{he2019s2gan}, we trained a ResNet-101 for each dataset as the age classifier to measure the age accuracy for different methods. Specifically, we translate the faces into the other three age groups, and we used the percentage of the faces falling into the exact target age group as the age accuracy, where the higher is the better. Table~\ref{tab:performance} presents the average age accuracy by different methods in each group. Our methods outperform other baseline methods by a large margin among all four age groups, demonstrating its capability of aging/rejuvenating faces with strong and natural aging effects.

\paragraph{\textbf{Identity Preservation}}\quad
We performed the face verification task to evaluate the identity preservation of different methods. Specifically, we used the pre-trained ResNet-101 face recognition model from ArcFace~\cite{deng2019arcface} to calculate the cosine similarity between the input and generated images. We reported the average cosine similarity of each pair of images as identity preservation. Here we emphasize that the verification rate is essentially computed by setting a threshold for the cosine similarity; that is, there is no difference between cosine similarity and face verification rate. The results are shown in Table~\ref{tab:performance}. Remarkably, our methods still achieve the highest scores at four age groups, confirming that our methods can achieve a satisfying age accuracy with the identities consistently preserved.

\paragraph{\textbf{Attribute Preservation}}\quad
Similar to age accuracy, we trained a classifier for each facial attribute of each dataset and reported the aging process starting from 30- as it is the hardest case in age progression/regression. Specifically, we evaluate the gender and race attributes for MORPH and only gender for CACD as there are almost all white people in CACD dataset. We argue that gender and race are the two most important facial attributes affecting face age synthesis. Other attributes such as hairstyle and expression are rarely discussed and not available in the benchmarked datasets. Table~\ref{tab:attribute_performance} presents the attribute preservation for different methods. Clearly, our method outperforms the baseline methods on the two datasets for different facial attributes. It is benefitting from the merit of invertible neural networks—-bijective mapping to preserve age-irrelevant information, which could best reduce the uncertainty of face transformations between two age groups. We note that the proposed method only needs the attribute labels during training compared to~\cite{liu2019attribute} that injects the attributes into the generator, therefore, it enjoys the flexibility of cGANs-based methods while consistently preserving facial attributes.

\subsection{Ablation Study}
To further validate the effectiveness of \methodname in age progression/regression, we performed an ablation study for the following variants: 1) GLOW: the trained GLOW model is used to manipulate the faces towards the target age group; 2) GLOW+attribute: we leverage the facial attributes to demonstrate that the prior attribute information can improve the attribute consistency during age progression/regression; 3) \methodname~w/o ICTM: we replace ICTM with a series of convolutional layers like cGANs to achieve the transformation in the latent space; and 4) \methodname: our proposed method distills the manipulation direction learned from the dataset to the ICTM. 

\begin{figure}[t!]
\centering
\includegraphics[width=1\linewidth]{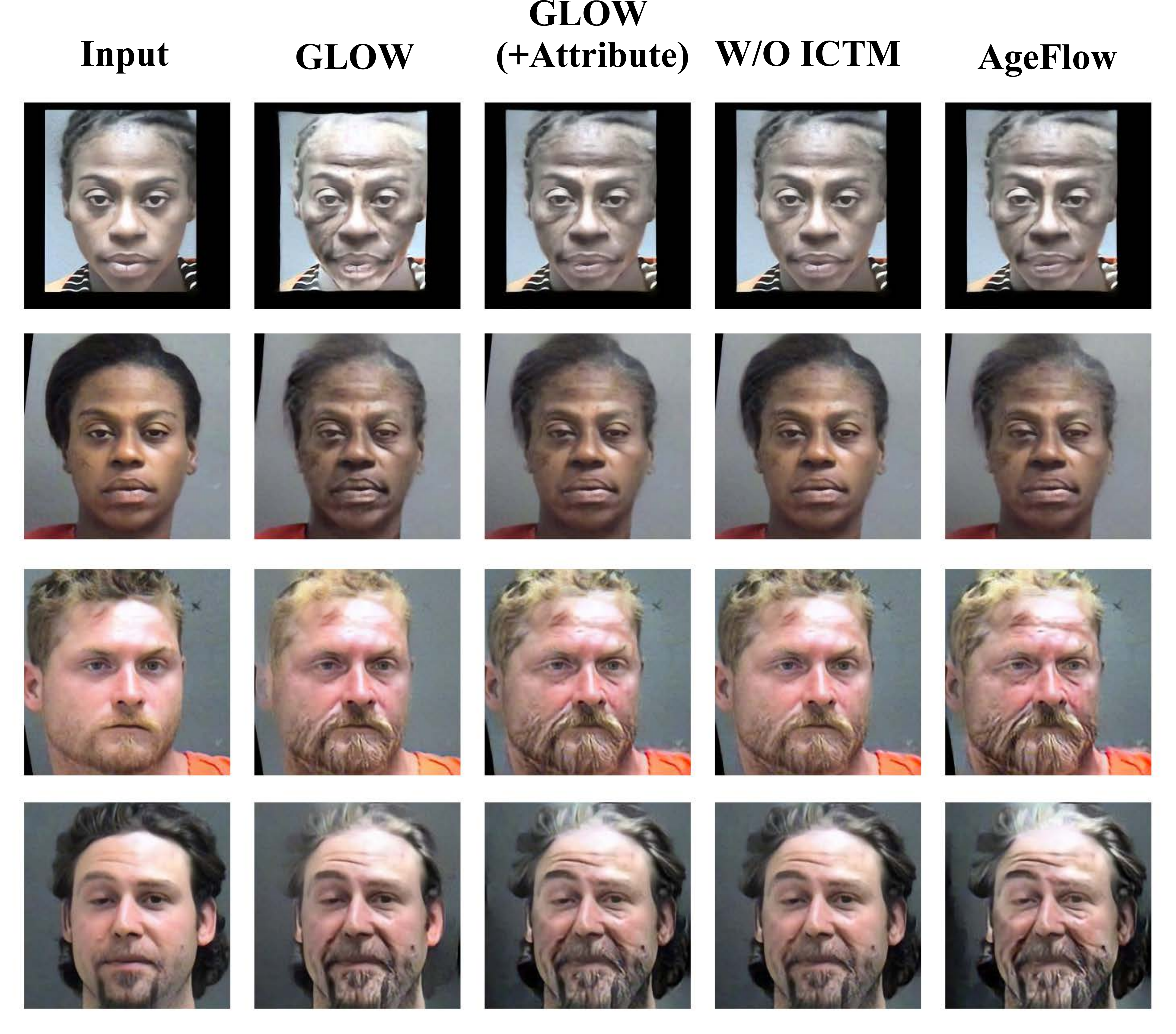}
\caption{Qualitative comparisons of different variants of \methodname.}
\label{fig:ablation}
\end{figure}

Table~\ref{tab:performance_ablation} shows the age accuracy and identity preservation while Table~\ref{tab:attribute_performance_ablation} presents the results of facial attribute preservation during age progression/regression. It can be observed that GLOW and its variant GLOW+attributes failed to achieve a satisfying age accuracy. This is because GLOW achieves age progression/regression by interpolating on the latent space, and they control the degree of age progression/regression by applying the manipulation direction. In a sense, different faces share the same patterns, which thus causes a low age accuracy. However, the attributes can be preserved by GLOW+attribute as it leverages the prior attributes information. Without ICTM, it fails to preserve the identity with the attributes changed, since the non-invertible introduces strong uncertainty to the age progression/regression process. 
Figure~\ref{fig:ablation} shows that the attributes were changed by the original GLOW but preserved by other variants. However, their aging/rejuvenation effects are not so strong as the proposed AgeFlow, since GLOW and GLOW+attribute are limited to the interpolations of latent space, and w/o ICTM cannot ensure an invertible between input and generated faces.
The results of the ablation study confirm the effectiveness of the prior attribution information and the proposed ICTM.  

\begin{table*}[ht]
  \centering
  \caption{Quantitative results for different variants of \methodname on MORPH and CACD datasets. The best results are shown in bold.}
  \begin{spacing}{1.0}
      \begin{tabular}{lrrrrclrrrr}
          \toprule
          \multicolumn{5}{c}{\textbf{MORPH}} && \multicolumn{5}{c}{\textbf{CACD}} \\ 
          \cmidrule{1-5} \cmidrule{7-11} 
          Age group & 30 - & 31 - 40 & 41 - 50 & 50 + && Age group & 30 - & 31 - 40 & 41 - 50 & 50 +\\
          \midrule
          \multicolumn{5}{c}{Age Accuracy~(\%)} & & \multicolumn{5}{c}{Age Accuracy~(\%)} \\
          \cmidrule{1-5} \cmidrule{7-11} 
          GLOW      & 47.92 & 52.75 & 60.46 & 65.23 & & GLOW & 49.92 & 49.64 & 52.61 & 32.13\\
          \quad+Attribute & 61.86 & 54.51 & 64.17 & 35.67 & & \quad+Attribute  & 67.35 & 42.39 & 55.39 & 46.95\\
          w/o ICTM      & 80.12 & 87.27 & 84.92 & 88.37 & & w/o ICTM       & 78.05 & 67.51 & 80.23 & 79.98\\
          AgeFlow   & \textbf{87.19}  & \textbf{91.85}  & \textbf{90.68}  & \textbf{90.71}  & & AgeFlow    & \textbf{86.49}  & \textbf{78.06}  & \textbf{85.91}  & \textbf{88.15}\\
          \cmidrule{1-5} \cmidrule{7-11} 
          \multicolumn{5}{c}{Cosine Similarity} & & \multicolumn{5}{c}{Cosine Similarity} \\
          \cmidrule{1-5} \cmidrule{7-11} 
          GLOW      & 0.809 & 0.772 & 0.912 & 0.721& & GLOW & 0.811 & 0.865 & 0.871 & 0.773\\
          \quad+Attribute  & 0.853 & 0.917 & \textbf{0.915} & 0.752& & \quad+Attribute & 0.895 & 0.910 & \textbf{0.925} & 0.796\\
          w/o ICTM      & 0.870 & 0.889 & 0.873 & 0.739& & w/o ICTM & 0.877 & 0.887 & 0.901 & 0.749\\
          AgeFlow   & \textbf{0.897} & \textbf{0.923} & 0.903 & \textbf{0.767}& & AgeFlow & \textbf{0.905} & \textbf{0.926} & 0.919 & \textbf{0.847}\\ 
          \bottomrule
          \end{tabular}
  \end{spacing}
  \label{tab:performance_ablation}
\end{table*}

\begin{table*}[t!]
  \centering
  \caption{Attribute preservation for different variants of \methodname on MORPH and CACD dataset. The best results are shown in bold.}
  \begin{tabular}{lcrrcrrrlcrr}
  \toprule
  & \multicolumn{7}{c}{\textbf{Gender}} && \multicolumn{3}{c}{\textbf{Race}} \\ 
  \cmidrule{2-8} \cmidrule{10-12} 
  & \multicolumn{3}{c}{MORPH} & \multicolumn{1}{l}{} & \multicolumn{3}{c}{CACD}  && \multicolumn{3}{c}{MORPH} \\
  \cmidrule{2-4} \cmidrule{6-8} \cmidrule{10-12} 
  Age Group & 31 - 40 & 41 - 50 & 51 +  & & 31 - 40 & 41 - 50 & 51 +  & & 31 - 40 & 41 - 50 & 51 + \\ 
  \midrule
  GLOW     & 95.02 & 94.89 & 94.60 & & 94.38 & 95.77 & 94.36 & & 97.34 & 96.88 & 96.45\\ 
  \quad+Attribute     & 97.62 & 98.95 & \textbf{98.22} & & \textbf{98.68} & 98.01 & \textbf{98.63} & & 98.72 & \textbf{98.83} & 97.09\\ 
  w/o ICTM     & 96.15 & 98.48 & 96.20 & & 98.14 & 96.66 & 98.30 & & 99.41 & 97.71 & 96.75\\ 
  AgeFlow     & \textbf{98.19} & \textbf{99.26} & 98.21 & & 98.63 & \textbf{98.02} & 98.49 & & \textbf{98.85} & 98.80 & \textbf{97.72}\\ 
  \bottomrule
  \end{tabular}
  \label{tab:attribute_performance_ablation}
\end{table*}

%% file: conc.tex
\section{Conclusion}\label{sec:conc}

In this paper, we proposed a novel framework, called \methodname, to integrate the advantages of both flow-based models and GANs, which enables age progression and regression to be invertible and achieves face transformation in the latent space with GANs and attribute-aware knowledge distillation loss. Extensive experiments demonstrate the superior performance over recently proposed baseline methods. Our work could shed new light on training invertible neural networks with high-resolution images and using GANs in the latent space learned from flow-based methods.

%% file: appendix.tex
\clearpage
\onecolumn
\appendix
\renewcommand \thepart{}
\renewcommand \partname{}
\part{\hfill \textsc{Appendix} \hfill}
\addcontentsline{toc}{section}{Appendix}
\numberwithin{equation}{section}
\setcounter{figure}{0}
\setcounter{table}{0}
\renewcommand\thetable{A.\arabic{table}}
\renewcommand\thefigure{A.\arabic{figure}}

\begin{figure*}[h]
\centering
\includegraphics[width=0.5\linewidth]{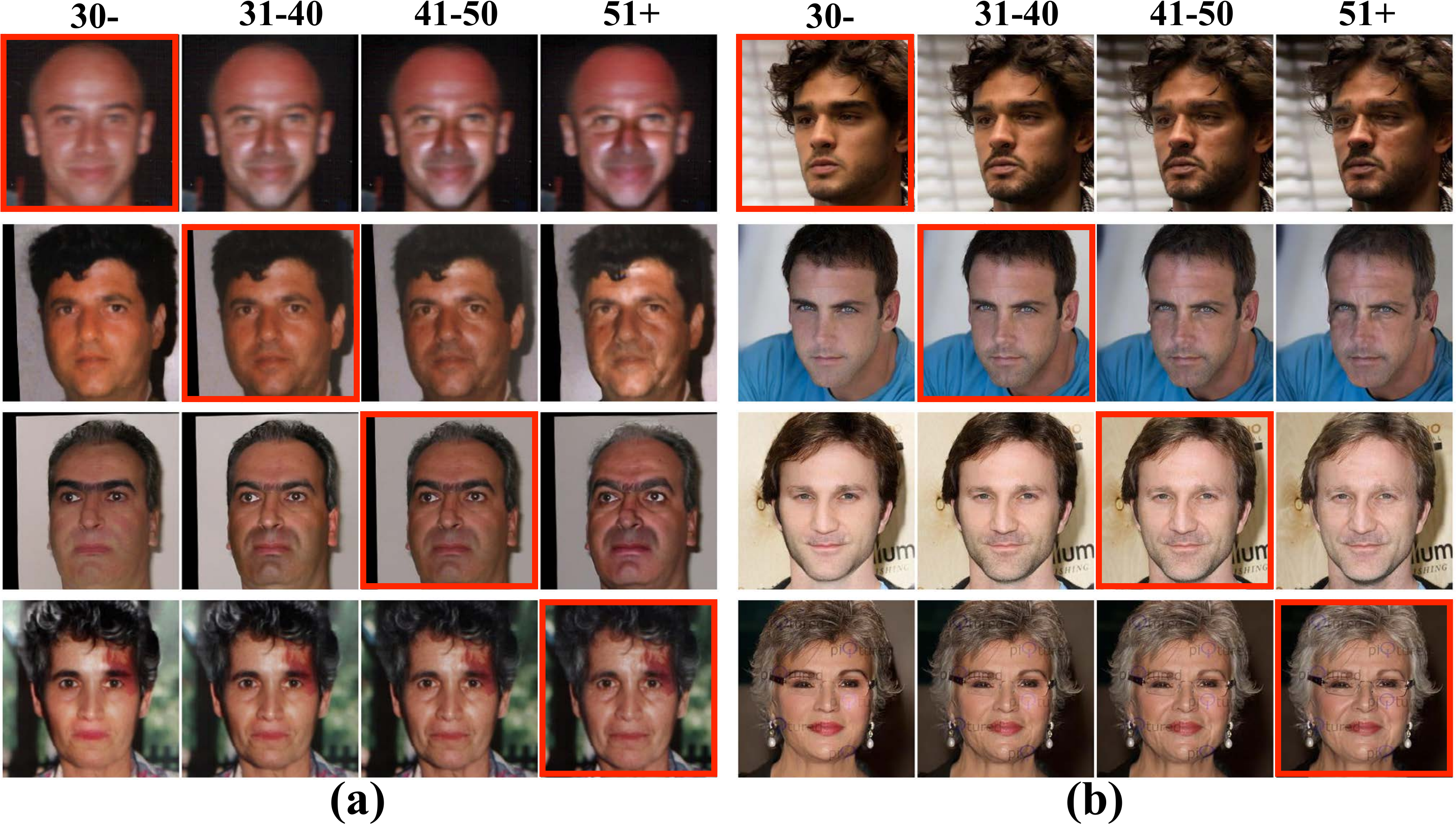}
\caption{The generated faces by AgeFlow on (a) FG-NET and (b) CelebA. Red boxes indicate input faces for progression/regression.}
\label{fig:generalization_ability}
\end{figure*}

\begin{figure*}[h]
\centering
\includegraphics[width=0.5\linewidth]{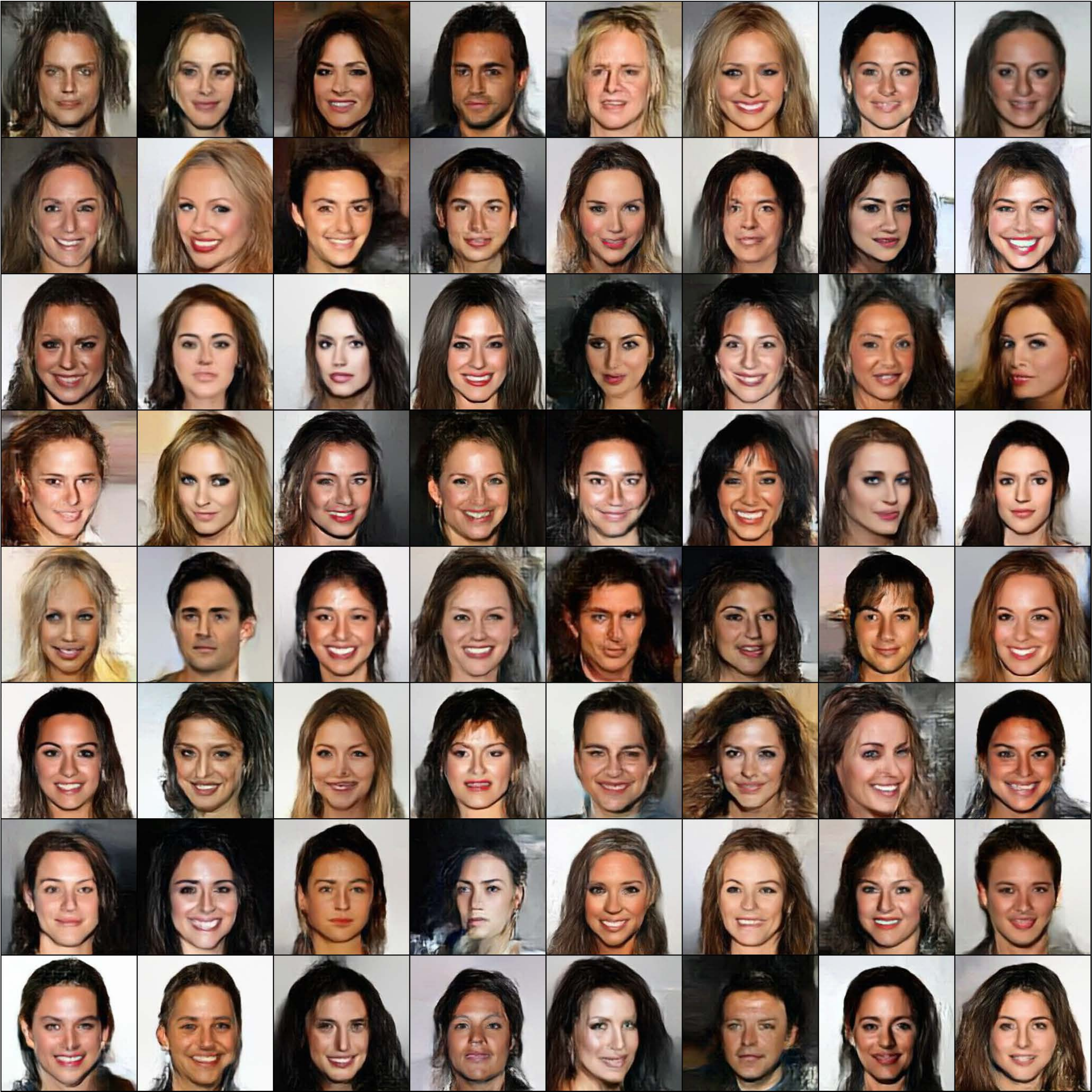}
\caption{The generated faces by the trained GLOW model with the size of $256\times 256$.}
\label{fig:glow}
\end{figure*}

Here we highlight that the pre-trained model would be very helpful for the research community. Figure~\ref{fig:generalization_ability} presents the sample results that apply the age progression/regression model trained on CACD~\cite{chen2015face} dataset to FG-NET and CelebA~\cite{liu2015faceattributes}, demonstrating the strong generalization ability of the proposed AgeFlow. Figure~\ref{fig:glow} shows our reproduced results of GLOW~\cite{kingma2018glow} with the image size of $256\times 256$ trained on CelebA~\cite{liu2015faceattributes}.


